\documentclass{article}

\usepackage{arxiv}

\usepackage[utf8]{inputenc} 
\usepackage[T1]{fontenc}    
\usepackage{hyperref}       
\usepackage{url}            
\usepackage{booktabs}       
\usepackage{amsfonts}       
\usepackage{nicefrac}       
\usepackage{microtype}      
\usepackage{lipsum}
\usepackage[ruled]{algorithm2e}
\usepackage{xcolor}
\usepackage{rotating}
\usepackage{bm}
\usepackage{amsmath,amssymb}
\usepackage{subcaption}
\newcommand*{\qed}{\null\nobreak\hfill\ensuremath{\square}}
\newenvironment{alg}[1][htb]
{
	\begin{algorithm}[#1]%
	}{\end{algorithm}}	
\newcommand{\bftab}{\fontseries{b}\selectfont}
\title{Walk Extraction Strategies for Node Embeddings with RDF2Vec in Knowledge Graphs}

\author{
  Gilles Vandewiele \\
  IDLab\\
  Ghent University -- imec\\
  Belgium \\
  \texttt{gilles.vandewiele@ugent.be} \\
  \And
  Bram Steenwinckel \\
  IDLab\\
  Ghent University -- imec\\
  Belgium \\
  \texttt{bram.steenwinckel@ugent.be} \\
  \And
  Pieter Bonte \\
  IDLab\\
  Ghent University -- imec\\
  Belgium \\
  \texttt{pieter.bonte@ugent.be} \\
  \And
  Michael Weyns \\
  IDLab\\
  Ghent University -- imec\\
  Belgium \\
  \texttt{michael.weyns@ugent.be} \\
  \And
  Heiko Paulheim \\
  Data and Web Science Group \\
  University of Mannheim \\
  Germany \\
  \texttt{heiko@informatik.uni-mannheim.de} \\
  \And
  Petar Ristoski \\
  IBM Research Almaden\\
  IBM\\
  United States of America \\
  \texttt{petar.ristoski@ibm.com} \\
  \And
  Filip De Turck \\
  IDLab\\
  Ghent University -- imec\\
  Ghent, Belgium \\
  \texttt{filip.deturck@ugent.be} \\
  \And
  Femke Ongenae \\
  IDLab\\
  Ghent University -- imec\\
  Ghent, Belgium \\
  \texttt{femke.ongenae@ugent.be} \\
}

\begin{document}
\maketitle

\begin{abstract}
	As KGs are symbolic constructs, specialized techniques have to be applied in order to make them compatible with data mining techniques. RDF2Vec is an unsupervised technique that can create task-agnostic numerical representations of the nodes in a KG by extending successful language modelling techniques. The original work proposed the Weisfeiler-Lehman (WL) kernel to improve the quality of the representations. However, in this work, we show both formally and empirically that the WL kernel does little to improve walk embeddings in the context of a single KG. As an alternative to the WL kernel, we propose five different strategies to extract information complementary to basic random walks. We compare these walks on several benchmark datasets to show that the \emph{n-gram} strategy performs best on average on node classification tasks and that tuning the walk strategy can result in improved predictive performances.
\end{abstract}

\keywords{Knowledge graphs \and Embeddings \and Representation Learning}
\section{Introduction}

As a result of the recent data deluge, we are increasingly confronted with more information than we can meaningfully make sense of. Above all, this information is characterised by contextual heterogeneity: its origins are semantically and syntactically diverse. Insights derived through traditional data mining procedures will be constrained precisely because such procedures fail to account for this staggering diversity. To deal with such a myriad of contextual environments and backgrounds, the Semantic Web's (SW) Linked Open Data (LOD) initiative can be used to interlink various data sources and unite them under a common queryable interface. The product of such a consolidation effort is often called a Knowledge Graph (KG). In addition to unifying information from various sources, KGs are able to enrich classical data formats by explicitly encoding relations between different data points in the form of edges. \\

Using KGs to enhance traditional data mining techniques with background knowledge is a relatively recent endeavour~\cite{Wilcke2017}. Because KGs are symbolic constructs, their compatibility with such techniques is rather limited. In fact, data mining techniques usually require inputs to be presented as numerical feature vectors, and are therefore unable to process background knowledge directly. With this in mind, some of the earliest knowledge-enhanced data mining approaches proceeded by extracting custom features from specific and generic relations inside the graph~\cite{ristoski2014comparison}. While these approaches produce human-interpretable variables, they have to be tailored to the task at hand and therefore require extensive effort. As an alternative to feature-based approaches, techniques can be applied to learn vector representations, called embeddings, for each of the entities inside a graph based on a limited set of global latent features~\cite{nickel2015review,Hamilton2017a}. These techniques are task-agnostic, which allows them to be used for different downstream tasks, such as predicting missing links inside a graph or categorizing different nodes~\cite{ristoski2016semantic}. \\

Natural language and graphs often share similarities. As an example, frequencies of language symbols or graph structures both tend to approximate Zipf's law. Techniques such as DeepWalk~\cite{perozzi2014deepwalk} and Node2Vec~\cite{grover2016node2vec} were among the first to leverage these similarities, by extending successful language modelling techniques, such as Word2Vec~\cite{mikolov2013distributed}, to deal with graph-based data. Their proposed techniques rely on the extraction of sequences of graph vertices, which are then fed as sentences to language models. Similarly, work on (deep) graph kernels also relies on language modelling to learn the latent representations of graph substructures \cite{vishwanathan2010graph,yanardag2015deep,kriege2020survey}. RDF2Vec is a technique that builds on the progress made by these previous two types of techniques by adapting random walks and the Weisfeiler-Lehman (WL) subtree kernel to directed graphs with labelled edges, i.e. KGs~\cite{ristoski2019rdf2vec}.  \\

In this work, we show that the WL kernel, while effective for measuring similarities between nodes or when working with regular graphs, offers little improvements in the context of a single KG with respect to walk embeddings. In response to this observation, we propose various alternative walk strategies for RDF data to improve upon basic random walks and compare them on different benchmark datasets. \\

The remainder of this paper is structured as follows. In Section \ref{sec:background}, background information is provided on KGs, RDF2Vec, walk embeddings and the WL kernel. Next, in Section \ref{sec:wl}, we provide a formal discussion of our claim with respect to the feasibility of Weisfeiler-Lehman subtree RDF graph kernels. Following this, Section \ref{sec:strategies} discusses a number of possible alternative walk strategies, including pseudo-code listings for each algorithm. Section \ref{sec:res} then describes the datasets used to evaluate these alternative strategies and lists the corresponding results. These results are subsequently discussed in Section \ref{sec:discussion}. Finally, in Section \ref{sec:conclusion} we conclude this work with a general reflection.

\section{Background}\label{sec:background}
\subsection{Knowledge graphs}\label{subsec:kg}

A KG is a multi-relational directed graph, $\mathbb{G} = (\mathbb{V},\ \mathbb{E},\ \ell)$, where $\mathbb{V}$ are the vertices in our graph, $\mathbb{E}$ the edges or predicates and $\ell$ a labelling function that maps each vertex or edge onto its corresponding label. It should further be noted that we can identify three different types of vertices: (i) \emph{entities}, (ii) \emph{blank nodes}, and (iii) \emph{literals}. We can simplify further analysis by applying a transformation to the knowledge graph which removes the multi-relational aspect, as done by de Vries et al.~\cite{de2015substructure}. This is done by representing each \texttt{(subject, predicate, object)} triple from the original knowledge graph by three labelled nodes and two unlabelled edges (\texttt{subject} $\rightarrow$ \texttt{predicate} and \texttt{predicate} $\rightarrow$ \texttt{object}).

\subsection{RDF2Vec}\label{subsec:rdf2vec}

Machine learning algorithms cannot work directly with graph-based data, as they require numerical vectors as input. RDF2Vec~\cite{ristoski2019rdf2vec} is an unsupervised, task-agnostic approach that solves this problem by first transforming the information of the nodes in the graph into numerical data, which are called latent representations or embeddings. The goal is to capture as much of the semantics as possible in the numerical representation, i.e. entities that are semantically related should be close to each other in the embedded space. RDF2Vec builds on word embedding techniques, which have shown great success in the domain of natural language processing. These word embedding techniques take a corpus of sentences as input, and learn a latent representation for each of the unique words within the corpus. Learning this latent representation can be done, for example, by learning to predict a word based on its context (continuous bag-of-words) or predicting the context based on a target word (skip-gram)~\cite{goldberg2014word2vec,mikolov2013distributed}. \\

\subsection{(Random) walk embeddings}
In the context of (knowledge) graphs, we can construct an input corpus by extracting walks. A walk is a sequence of vertices that can be found in the graph by traversing the directed links. We can notate a walk of length n as follows:
\begin{equation}
v_0 \rightarrow v_1 \rightarrow \ldots \rightarrow v_{n-1}
\end{equation}
We can then use the graph labelling function to create a sentence:
\begin{equation}
\ell(v_0) \rightarrow \ell(v_1) \rightarrow \ldots \rightarrow \ell(v_{n-1})
\end{equation}
It should be noted that, due to the previously discussed transformation of our KG, nodes with an even index in the walk correspond to entities of the original knowledge graph, while nodes with an odd index correspond to predicates. \\

The most straightforward strategy to extract walks is by doing a breadth-first traversal of the graph starting from the nodes of interest. Since the total number of walks that can be extracted grows exponentially in function of the depth, sampling can be applied after each iteration of breadth-first traversal. This sampling can either be guided by some metric, resulting in a collection of biased walks~\cite{cochez2017biased}, or can be performed at random, which results in \emph{random} walks.

\subsection{Weisfeiler-Lehman kernel}
The WL kernel was proposed as an extension to the labelling function. The WL kernel is an algorithm to test whether two graphs were isomorphic in polynomial time~\cite{weisfeiler1968reduction}. The intuition behind the algorithm was to assign new labels to each of the nodes, where each of the newly assigned labels captured the information of an entire subgraph up to a certain depth. This algorithm was later adapted to serve as a kernel, or similarity measure, between graphs~\cite{de2013fast,shervashidze2011weisfeiler}, by counting the number of WL labels two graphs had in common. The WL relabelling for a node $v$ is performed by recursively hashing the concatenation of the label of an entity and the (sorted) labels the nodes of its neighbours:
\begin{align} \label{eq:wl}
\scriptsize
\textsc{wl}_{0}(v) =&\ \scriptsize \ell(v) \nonumber \\ 
\scriptsize
\textsc{wl}_{k}(v) =&\ \scriptsize \textsc{hash}(\textsc{wl}_{k - 1}(v) \oplus \textsc{sort}(\{\textsc{wl}_{k - 1}(v_n)\ |\ v_n \in \bm{\mathcal{N}}(v)\}))
\end{align} 
with $k \geq 1$, $\textsc{wl}_{k}(v)$ the corresponding \textsc{wl} label of a node after $k$ iterations, $\textsc{hash}$ a hashing function, $\textsc{sort}$ a sorting function, $\oplus$ the string concatenation operator and $\bm{\mathcal{N}}(v)$ the neighbourhood of $v$. We assume that the hashing function does not produce collisions. This can easily be achieved by, for example, mapping each unique label to an integer. The neighbourhood of a vertex $v$ is defined as the set of vertices $v'$ with an edge going to $v$:
\begin{equation} \label{eq:neighbourhood}
\bm{\mathcal{N}}(v) = \{v'\ |\ (v', v) \in \mathbb{E}\}
\end{equation}

\section{Weisfeiler-Lehman kernel for Knowledge Graphs}\label{sec:wl}
Ristoski et al. proposed to use the \textsc{wl} kernel in order to relabel nodes as an alternative to extracting random walks~\cite{ristoski2019rdf2vec}. We will refer to this as the \emph{WL} strategy in the remainder of this paper. However, we argue that this \emph{WL} strategy provides no additional information with respect to entity representations when extracting a fixed number of random walks from a knowledge graph. We now formally demonstrate this claim on al three types of vertices:

\subsection{Entities}
The \emph{wl} strategy brings little to no added value in comparison to random walks, when applied to entities in a knowledge graph, due to the properties of \textsc{rdf} data. Entities in \textsc{rdf} are represented by Uniform Resource Identifiers (\textsc{uri}), which need to be unique\footnote{\url{https://www.w3.org/DesignIssues/Axioms.html}}. As such:
\begin{equation} \label{eq:rdf}
\ell(x) = \ell(y) \iff x = y
\end{equation}

Due to this property, \textsc{wl} relabelling, when applied on \textsc{rdf} data, is nothing more than a bijection from the hops in random walks to the hops in the walks obtained through \textsc{wl} relabelling. This means that \textsc{wl} relabelling does not add any useful additional information.  When two \textsc{wl} labels are equal, their underlying entities are always equal as well. We will now prove this formally. First, from Eq.~\ref{eq:wl}  we can deduce that when two \textsc{wl} labels of two nodes are equal, then at least the labels of these nodes should be equal and they should have the same neighbours\footnote{Proof omitted due to space restrictions, but as \textsc{wl} is recursive, it can be proven through induction.}. Formally this means that:
\begin{equation} \label{eq:defN}
\scriptsize
\textsc{wl}_{k}(v_j)=\textsc{wl}_{k}(v_i) \iff \ell(v_j)=\ell(v_i) \wedge \bm{\mathcal{N}}(v_j) = \bm{\mathcal{N}}(v_i)
\end{equation}

\subsubsection*{Proof:} \small $\textsc{wl}_{k}(v_j)=\textsc{wl}_{k}(v_i) \iff v_j = v_i$

\par
\textbf{\normalsize Step 1:} $\textsc{wl}_{k}(v_j)=\textsc{wl}_{k}(v_i) \implies v_j = v_i$
\begin{flalign*}
\textsc{wl}_{k}(v_j) = \textsc{wl}_{k}(v_i) & \implies \ell(v_j) = \ell(v_i) \wedge \bm{\mathcal{N}}(v_j) = \bm{\mathcal{N}}(v_i) & \text{(Eq.~\ref{eq:defN})}\\
&\implies \ell(v_j) = \ell(v_i) & \text{(Conjunction Elim.)}  \\
&\implies v_j = v_i & \text{(Eq.~\ref{eq:rdf})} \\ 
& & \hfill \qed
\end{flalign*}

\textbf{\normalsize Step 2:} $v_j = v_i \implies \textsc{wl}_{k}(v_j)=\textsc{wl}_{k}(v_i)$
\begin{flalign*}
v_j = v_i & \implies \ell(v_j) = \ell(v_i)  & \text{(Eq.~\ref{eq:rdf})}\\
&\implies \ell(v_j) = \ell(v_i) \wedge \bm{\mathcal{N}}(v_j) = \bm{\mathcal{N}}(v_i) & \text{(Eq.~\ref{eq:neighbourhood} and $v_j = v_i$)}\\
&\implies \textsc{wl}_{k}(v_j)=\textsc{wl}_{k}(v_i) & \text{(Eq.~\ref{eq:defN})} \\
& & \hfill \qed
\end{flalign*}

\normalsize

\subsection{Blank nodes and literals} 
In contrast to \emph{entities}, Eq.~\ref{eq:rdf} does not hold for blank nodes and literals. This implies that multiple nodes could have the same original label but have different WL labels (one-to-many mapping). However, the added value of WL is limited even in these cases due to the fact that blank nodes rarely have the exact same neighbourhoods and because literals only have one incoming edge and no outgoing edges. Moreover, due to the fact that RDF2Vec treats each hop in the walk as categorical data, RDF2Vec does not handle literals well.

\section{Custom walk extraction strategies}\label{sec:strategies}

Based on the observation probably discussed, we now identify two types of strategies to construct a corpus of walks:
\begin{description}
	\item[Type 1 - Extraction:] strategies that define how walks for each of the entities are extracted. The \emph{random} walk strategy is an example of such a strategy, where breadth-first traversal is applied to extract walks. 
	\item[Type 2 - Transformation:] strategies that transform walks extracted by a \textbf{Type 1} strategy. The \emph{WL} strategy is an example of this type. In order for such a strategy to provide information complementary to the originally provided walks, it must define a \emph{one-to-many} or \emph{many-to-one} mapping from the original labels to the new labels.
\end{description}

We now propose five different strategies complementary to the \emph{random} strategy. One of these strategies can be classified as being of \textbf{Type 1} while the other four are of \textbf{Type 2}.

\subsection{Community hops}
	As opposed to iteratively extending the walk with neighbours of a vertex, we could allow with a certain probability for teleportation to a node that has properties similar to a certain neighbor~\cite{keikha2018community}. In order to group nodes with similar properties together, unsupervised community detection can be applied~\cite{fortunato2010community}. In this work, we opted to use the Louvain method~\cite{blondel2008fast} due to its excellent trade-off between speed and clustering quality. The idea of introducing community hops is to capture implicit relations between nodes that are not explicitly modelled in the KG, and to allow for including related pieces of knowledge in the walks which are otherwise out of reach. We provide pseudo-code for this strategy in Algorithm~\ref{alg:community_walk}. This strategy is of \textbf{Type 1}. We will refer to this strategy as \emph{community}.
	
	\begin{alg}[h!]
		\small
		\SetStartEndCondition{ }{}{}%
		\SetKwProg{Fn}{def}{\string:}{}
		\SetInd{1em}{0.5em}
		\SetKwFunction{Range}{range}
		\SetKw{KwTo}{in}\SetKwFor{For}{for}{\string:}{}%
		\SetKwIF{If}{ElseIf}{Else}{if}{:}{elif}{else:}{}%
		\SetKwFor{While}{while}{:}{fintq}%
		\SetKwInOut{Input}{Input}
		\SetKwInOut{Output}{Output}
		\SetKw{In}{in}
		\SetKw{Or}{or}
		\SetKw{Not}{not}
		\SetKwComment{tcp}{\# }{}%
		\DontPrintSemicolon
		\AlgoDontDisplayBlockMarkers\SetAlgoNoEnd\SetAlgoNoLine%
		
		\SetAlgoLined
		
		\tcp{List of communities and dictionary \{vertex: community\}}
		com, com\_map = \textsc{\texttt{com\_detection}}($\mathbb{G}$)
		
		walks = \{ (v,) \}
		
		\For{\upshape $d$ \In \textsc{\texttt{range}}($depth$)}{
			new = set()
			
			\For{\upshape walk \In walks}{
				\For{\upshape n \In \textsc{\texttt{get\_neighbours}}($\mathbb{G}$, v)}{
					\tcp{Sample neighbourhood}
					\If{\upshape \textsc{\texttt{random}}() $<$ p}{
						new.\textsc{\texttt{add}}(walk + (n,))
					}
					
					\;
					
					\tcp{Hop to community}
					\If{\upshape \textsc{\texttt{random}}() $<$ hop\_prob}{
						c\_n = com[com\_map[n]]
						
						hop = \textsc{\texttt{choice}}(c\_n)
						
						new.\textsc{\texttt{add}}(walk + (hop,))
					}
					
				}
			}
			
			walks = new
		}
		
		\Return walks
		
		\caption{\small \textsc{\texttt{community\_walk}}($\mathbb{G}$, v, depth, p, hop\_prob)}
		\label{alg:community_walk}
	\end{alg}

\subsection{Anonymous walks}
	The random walks discussed in the previous section can be anonymized, which transforms label information into positional information. More formally, a walk $w = v_0 \rightarrow v_1 \rightarrow \ldots \rightarrow v_n$, is transformed into $f(v_0) \rightarrow f(v_1) \rightarrow \ldots \rightarrow f(v_n)$ with $f(v_i) = \min(\{i\ |\ w[i]\ =\ v_i\})$, which corresponds to the first index where $v_i$ can be found in the walk $w$~\cite{ivanov2018anonymous}. The notion behind anonmyous walks is that local graph structures often bear enough information for encoding and reconstructing a graph, even when ignoring the node labels, i.e., the mere topology surrounding a node is often sufficient for identifying that node. Ignoring the labels, on the other hand, allows for a computationally efficient generation of the walks. We present pseudo-code for this transformation in Algorithm~\ref{alg:anonymize_walks}. This strategy is of \textbf{Type 2}. We will refer to this strategy as \emph{anonymous}.
	
	\begin{alg}[h!]
		\small
		\SetStartEndCondition{ }{}{}%
		\SetInd{1em}{0.5em}
		\SetKwProg{Fn}{def}{\string:}{}
		\SetKwFunction{Range}{range}
		\SetKw{KwTo}{in}\SetKwFor{For}{for}{\string:}{}%
		\SetKwIF{If}{ElseIf}{Else}{if}{:}{elif}{else:}{}%
		\SetKwFor{While}{while}{:}{fintq}%
		\SetKwInOut{Input}{Input}
		\SetKwInOut{Output}{Output}
		\SetKw{In}{in}
		\SetKw{Or}{or}
		\SetKw{Not}{not}
		\SetKwComment{tcp}{\# }{}%
		\DontPrintSemicolon
		\AlgoDontDisplayBlockMarkers\SetAlgoNoEnd\SetAlgoNoLine%
		
		\SetAlgoLined
		
		anon\_walks = [ ]
		
		\For{\upshape walk \In walks}{
			new = [ walk[0] ]
			
			\For{\upshape hop \In walk[1:]}{
				new.\textsc{\texttt{append}}(walk.\textsc{\texttt{index}}(hop))
			}
			
			anon\_walks.\textsc{\texttt{append}}(new)
		}
		
		\Return anon\_walks
		
		\caption{\small \textsc{\texttt{anonymize}}(walks)}
		\label{alg:anonymize_walks}
	\end{alg}
	
\subsection{Walkets}
	Walks can be transformed into walklets, which are walks of length two consisting of the root of the original walk and one of the hops. Provided a walk $w = v_0 \rightarrow v_1 \rightarrow \ldots \rightarrow v_n$, we can construct sets of walklets $\{(v_0, v_{i})\ |\ 1 \leq i \leq n \}$~\cite{perozzi2017don}. While standard RDF2Vec does not consider the distance between two nodes in a walk, walklets are explicitly created for different scales. Hence, they allow for such a distinction between a direct neighbor and a node which is further away. Pseudocode for this approach is provided in Algorithm~\ref{alg:walklets}. This strategy is of \textbf{Type 2}. We will refer to this strategy as \emph{walklet}.
	
	\begin{alg}[h!]
		\small
		\SetStartEndCondition{ }{}{}%
		\SetInd{1em}{0.5em}
		\SetKwProg{Fn}{def}{\string:}{}
		\SetKwFunction{Range}{range}
		\SetKw{KwTo}{in}\SetKwFor{For}{for}{\string:}{}%
		\SetKwIF{If}{ElseIf}{Else}{if}{:}{elif}{else:}{}%
		\SetKwFor{While}{while}{:}{fintq}%
		\SetKwInOut{Input}{Input}
		\SetKwInOut{Output}{Output}
		\SetKw{In}{in}
		\SetKw{Or}{or}
		\SetKw{Not}{not}
		\SetKwComment{tcp}{\# }{}%
		\DontPrintSemicolon
		\AlgoDontDisplayBlockMarkers\SetAlgoNoEnd\SetAlgoNoLine%
		
		\SetAlgoLined
		
		walklets = set()
		
		\For{\upshape walk \In walks}{
			\For{\upshape i \In \textsc{\texttt{range}}(1, $|$walk$|$)}{
				walklets.\textsc{\texttt{add}}((walk[0], walk[i]))
			}
		}
		
		\Return walklets
		
		\caption{\small \textsc{\texttt{walklets}}(walks)}
		\label{alg:walklets}
	\end{alg}

\subsection{Hierarchical random walk (HALK)}
	The frequency of entities in a knowledge graph often follows a long-tailed distribution, similar to natural language. Entities rarely occurring often carry little information, and increase the number of hops between the root and potentially more interesting entities. As such, the removal of rare entities from the random walks can increase the quality of the generated embeddings while decreasing the memory usage~\cite{schlotterer2019investigating}. Pseudo-code for this strategy is provided in Algorithm~\ref{alg:halk}. This strategy is of \textbf{Type 2}. We will refer to this strategy as \emph{HALK}.
	
	\begin{alg}[h!]
		\small
		\SetStartEndCondition{ }{}{}%
		\SetKwProg{Fn}{def}{\string:}{}
		\SetInd{1em}{0.5em}
		\SetKwFunction{Range}{range}
		\SetKw{KwTo}{in}\SetKwFor{For}{for}{\string:}{}%
		\SetKwIF{If}{ElseIf}{Else}{if}{:}{elif}{else:}{}%
		\SetKwFor{While}{while}{:}{fintq}%
		\SetKwInOut{Input}{Input}
		\SetKwInOut{Output}{Output}
		\SetKw{In}{in}
		\SetKw{Or}{or}
		\SetKw{Not}{not}
		\SetKwComment{tcp}{\# }{}%
		\DontPrintSemicolon
		\AlgoDontDisplayBlockMarkers\SetAlgoNoEnd\SetAlgoNoLine%
		
		\SetAlgoLined

		\tcp{Count nr. of walks a hop occurs}
		
		counts = \{ \}
		
		\For{\upshape i \In \textsc{\texttt{range}}($|$walk$|$)}{
			\For{\upshape hop \In walks[i]}{
				\If{\upshape hop \Not \In frequencies}{
					counts[hop] = \{i\}
				}\Else{
					counts[hop].\textsc{\texttt{add}}(i)
				}
			}
		}
		
		\;
		
		\tcp{Skip rare hops}
		
		halk\_walks = [ ]
		
		\For{\upshape thresh in thresholds}{
			\For{\upshape walk in walks}{
				new = [ walk[0] ]
				
				\For{\upshape hop \In walk[1:]}{
					\If{\upshape $\frac{|\textrm{counts[hop]}|}{|\textrm{walks}|} \geq$ thresh}{
						new.\textsc{\texttt{append}}(hop)
					}
				}
				halk\_walks.\textsc{\texttt{append}}(new)

			}
		}
		
		\;
		
		\Return halk\_walks
		\caption{\small \textsc{\texttt{halk}}(walks, thresholds)}
		\label{alg:halk}
	\end{alg}
	
\subsection{N-Gram walks}
	Another approach that defines a one-to-many mapping is relabelling n-grams in the random walks. The intuition behind this is that the predecessors of a node that two different walks have in common can be different. Additionally, we can inject wildcards into the walk before relabelling n-grams~\cite{vandewiele2019inducing}. The injection of wildcards allows subsequences with small differences to be mapped onto the same label. Pseudo-code for this strategy is provided in Algorithm~\ref{alg:ngram}. This strategy is of \textbf{Type 2}. We will refer to this strategy as \emph{n-gram}.
	
	 \begin{alg}[h!]
	 	\small
		\SetStartEndCondition{ }{}{}%
		\SetInd{1em}{0.5em}
		\SetKwProg{Fn}{def}{\string:}{}
		\SetKwFunction{Range}{range}
		\SetKw{KwTo}{in}\SetKwFor{For}{for}{\string:}{}%
		\SetKwIF{If}{ElseIf}{Else}{if}{:}{elif}{else:}{}%
		\SetKwFor{While}{while}{:}{fintq}%
		\SetKwInOut{Input}{Input}
		\SetKwInOut{Output}{Output}
		\SetKw{In}{in}
		\SetKw{Or}{or}
		\SetKw{Not}{not}
		\SetKwComment{tcp}{\# }{}%
		\DontPrintSemicolon
		\AlgoDontDisplayBlockMarkers\SetAlgoNoEnd\SetAlgoNoLine%
		
		\SetAlgoLined
		
		\tcp{Introduce wildcards in the walks}
		
		extended\_walks = walks
		
		\For{\upshape walk \In walks}{
			
			idx = \textsc{\texttt{range}}(1, $|$walk$|$)
			
			combs = \textsc{\texttt{combinations}}(idx, n\_wild)
			
			\For{\upshape comb \In combs}{
				
				new = walk
				
				\For{\upshape i \In comb}{
					new[i] = `*'
				}
				
				extended\_walks.\textsc{\texttt{append}}(new)
				
			}
			
		}
		
		\tcp{Relabel ngrams in the walk}
		
		ngram\_walks = [ ]
		
		map = \{ \}
		
		\For{\upshape walk \In extended\_walks}{
			
			new = walk[:n]
			
			\For{\upshape i \In \textsc{\texttt{range}}(n, $|$walk$| + 1$)}{
				ngram = walk[i-n:i]
				
				\If{\upshape ngram \Not \In map}{
					map[ngram] =  $|$map$|$
				}
				
				new.\textsc{\texttt{append}}(map[ngram])
				
			}
			
			ngram\_walks.\textsc{\texttt{append}}(new)
			
		}
		
		\Return ngram\_walks
		
		\caption{\small \textsc{\texttt{ngram}}(walks, n, n\_wild)}
		\label{alg:ngram}
	\end{alg}

\subsection{Example}

\begin{figure}
	\centering
	\includegraphics[scale=0.75]{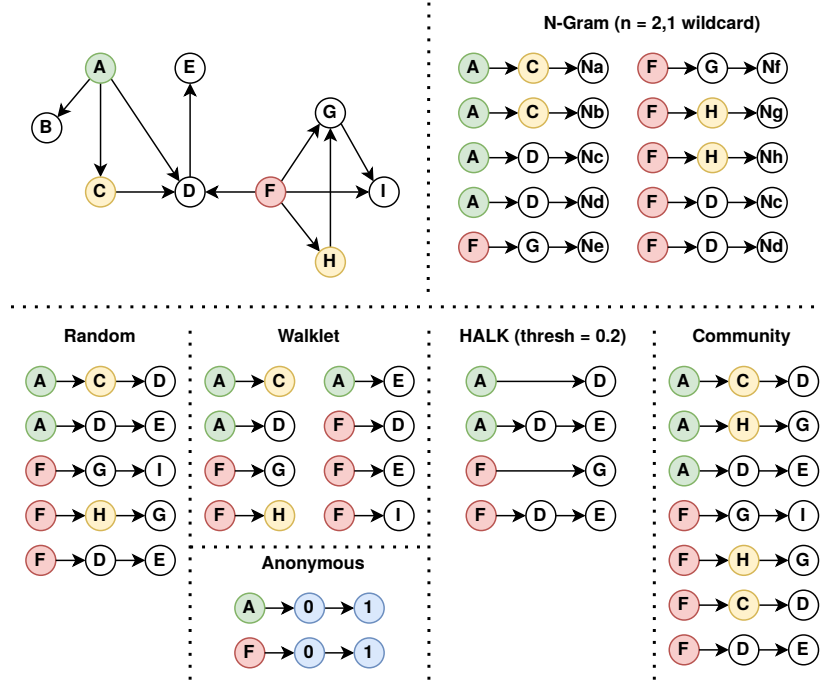}
	\caption{An example of each of our proposed strategies. We extract walks with the \emph{random} and \emph{community} strategy of exactly depth 2 from nodes ``A" and ``F". For other strategies, we transform the walks extracted by the \emph{random} strategy. Nodes ``C" and ``H" belong to the same community.}
	\label{fig:walk_example}
\end{figure}

In order to further clarify each of the proposed strategies, we provide an example in Figure~\ref{fig:walk_example}.

\section{Results}\label{sec:res}

To evaluate the impact of custom walking strategies, we measure the predictive performance on different datasets and various tasks.

\subsection{Datasets}

Three different types of datasets are used, in order to ensure enough variation in our evaluation. Moreover, these datasets are commonly used in (knowledge) graph-based machine learning studies.

\subsubsection{Node classification benchmark datasets} We will be using four benchmark data sets, each describing knowledge graphs, that serve as benchmarks for node classification and are available from a public repository set up by Ristoski et al.~\cite{ristoski2016collection}. The names of these benchmark datasets are AIFB, MUTAG, BGS and AM. For each of these data sets, we remove triples with specific predicates that would leak the target from our knowledge graph, as provided by the original authors. Moreover, a predefined split into train and test set, with the corresponding ground truth, is provided by the authors, which we used in our experiments.
\subsubsection{Citation networks} We converted three citation networks~\cite{sen2008collective}, which describe scientific papers, to knowledge graphs. The three citation networks used are CORA, CITESEER and PUBMED. Each paper is represented by a bag-of-word or tf-idf representation of their content and a list of citations to other papers in the network. A fixed train/test split is provided for each of the datasets and the associated task is to categorize each of the papers into the correct research domain, which can be seen as a node classification task. For each paper $p$, we obtained the words $w$ from the bag-of-words or tf-idf vector with a value greater than 0 and add the following triples to our KG: $\{ (p,\ hasWord,\ w)\ |\ f(p, w) > 0 \}$, with $f(p, w)$ a function that retrieves the bag-of-word or tf-idf value of word $w$ for paper $p$. Moreover, for each paper $p'$ cited by $p$ we add the following triple: $(p,\ cites,\ p')$.
\subsubsection{DBpedia} We use the English version of the 2016-10 DBpedia dataset \cite{lehmann2015dbpedia}, which contains 4,356,314 entities and 52,689,448 triples in total. In our evaluation, we only consider object properties, and ignore datatype properties and literals. We use the obtained embeddings in multiple different downstream tasks: 5 different classification tasks (AAUP, Cities, Forbes, Albums and Movies), document similarity and entity relatedness. For more details on each of these tasks, we refer the reader through to the original RDF2Vec paper by Ristoski et al.~\cite{ristoski2019rdf2vec}.

\subsection{Setup}

	For each of the entities in all of the datasets, walks of depth $4$ are exhaustively extracted. A depth of $4$ is chosen as it results in the best predictive performances on average for all strategies and datasets. Only for the entities of DBpedia, the maximum number of walks per entity is limited to 500. These walks are then provided to a Word2Vec model to create $500$-dimensional embeddings. The hyper-parameters of the Word2Vec model are the same for all experiments in this study. Skip-Gram is used, the window size is equal to $5$ and the maximum number of iterations is equal to $10$ with negative sampling set to $25$. These configurations are identical to the original RDF2Vec study. The embeddings are learned, in an unsupervised manner, for both the train and test set. For node classification tasks, embeddings are fed to a Support Vector Machine (SVM) classifier with Radial Basis Function (RBF) kernel. The regularization strength of the SVM is tuned to be one of $\{0.001, 0.01, 0.1, 1.0, 10.0, 100.0, 1000.0\}$. For tasks other than node classification, an evaluation framework is used~\cite{pellegrino2020configurable}. For document similarity, we measure the Pearson's linear correlation coefficient, Spearman's rank correlation and their harmonic mean. For entity relatedness, we measure the Kendall's rank correlation coefficient. For the benchmark datasets and citation networks, a pre-defined train/test split is used and experiments are repeated 5 times in order to report a corresponding standard deviation. For the tasks involving DBpedia data, 10-fold cross-validation is used and experiments are only repeated once for timing reasons. Moreover, the \textit{community} strategy was excluded from the DBpedia experiments, as it cannot be efficiently performed on large knowledge graphs. \\
	
	For each of the walking strategies, we tune the following hyper-parameters using either a provided validation set or by using cross-validation on the train set:
	\begin{itemize}
		\item The \emph{random}, \emph{anonymous} and \emph{walklet} strategies are hyper-parameter-free.
		\item For the \emph{n-gram} walker, we tune $n \in [1, 2, 3]$  and the number of introduced wildcards to be either 0 or 1.
		\item For the \emph{community} strategy, we set the resolution of the Louvain algorithm to 1.0~\cite{lambiotte2008laplacian} and the probability to teleport to a node from the community to 10\%.
		\item For the \emph{WL} strategy, we use the original algorithm used by Ristoski et al.~\cite{ristoski2019rdf2vec}. We set the number of iterations of the Weisfeiler-Lehman kernel to 4 and extract walks of fixed depth for each of the iterations, including zero. This causes the WL walker to extract 5 times as many walks as the random walker, which causes the results to differ from those of the random walk strategy.
		\item For the \emph{HALK} strategy, we extract sets of walks using different thresholds: $[0.0, 0.1, 0.05, 0.01, 0.005, 0.001, 0.0005, 0.0001]$.
	\end{itemize}

\subsection{Evaluation results}

	The results for the various classification tasks are provided in Table~\ref{tab:accuracies}. The results for the document similarity and entity relatedness task are provided in Table~\ref{tab:results2}.

\begin{table*}
	\centering
	\scriptsize
	\begin{tabular}{l|r|r||r|r|r|r|r}
		\toprule \multicolumn{1}{c}{} & \multicolumn{1}{c}{\textbf{Random}} & \multicolumn{1}{c}{\textbf{WL}} & \multicolumn{1}{c}{\textbf{Walkets}} & \multicolumn{1}{c}{\textbf{Anonymous}} & \multicolumn{1}{c}{\textbf{HALK}} & \multicolumn{1}{c}{\textbf{N-Gram}} & \multicolumn{1}{c}{\textbf{Community}} \\  \toprule \toprule
		
		AIFB & 86.11 $\pm$ 2.48 & \bftab 91.67 $\pm$ 0.00 & 63.89 $\pm$ 0.00 & 41.67 $\pm$ 0.00 & 86.11 $\pm$ 0.00 & 88.33 $\pm$ 1.11 & 88.89 $\pm$ 1.76  \\ 
		MUTAG & 76.76 $\pm$ 0.59 & 75.00 $\pm$ 2.46 & 72.06 $\pm$ 0.00 & 66.18 $\pm$ 0.00 & 75.00 $\pm$ 0.00 & \bftab 77.65 $\pm$ 2.85 & 74.71 $\pm$ 3.99  \\ 
		BGS & 79.31 $\pm$ 0.00 & 80.69 $\pm$ 6.40 & 65.52 $\pm$ 0.00 & 65.52 $\pm$ 0.00 & 80.00 $\pm$ 4.57 & 83.45 $\pm$ 4.02 & \bftab 84.14 $\pm$ 3.52  \\ 
		AM & 75.56 $\pm$ 2.70 & 82.53 $\pm$ 1.68 & 47.47 $\pm$ 0.00 & 34.85 $\pm$ 0.00 & 80.10 $\pm$ 0.88 & \bftab 84.44 $\pm$ 2.22 & 73.94 $\pm$ 2.70  \\ \midrule \midrule
		
		CORA & \bftab 77.20 $\pm$ 0.00 & 74.32 $\pm$ 1.56 & 58.20 $\pm$ 0.00 & 14.30 $\pm$ 0.00 & 76.62 $\pm$ 0.36 & 76.46 $\pm$ 0.78 & 67.92 $\pm$ 1.22  \\ 
		CITESEER & 64.68 $\pm$ 1.58 & 64.02 $\pm$ 1.46 & 38.40 $\pm$ 0.00 & 16.00 $\pm$ 0.00 & \bftab 66.90 $\pm$ 0.00 & 65.38 $\pm$ 1.22 & 58.66 $\pm$ 0.50  \\ 
		PUBMED & 75.66 $\pm$ 1.36 & 73.70 $\pm$ 2.87 & 68.30 $\pm$ 0.00 & 24.20 $\pm$ 0.00 & 75.56 $\pm$ 0.08 & \bftab 78.48 $\pm$ 0.35 & 54.64 $\pm$ 2.40  \\ \midrule \midrule
		
		DBP: AAUP & \multicolumn{1}{l|}{67.94}  & \multicolumn{1}{l|}{\bftab 69.88}  & \multicolumn{1}{l|}{69.27}  & \multicolumn{1}{l|}{54.73}  & \multicolumn{1}{l|}{60.08}  & \multicolumn{1}{l|}{66.96}  & \multicolumn{1}{c}{/} \\ 
		DBP: Cities & \multicolumn{1}{l|}{79.07}  & \multicolumn{1}{l|}{79.12}  & \multicolumn{1}{l|}{79.08}  & \multicolumn{1}{l|}{55.34}  & \multicolumn{1}{l|}{73.34}  & \multicolumn{1}{l|}{\bftab 79.79}  & \multicolumn{1}{c}{/} \\ 
		DBP: Forbes & \multicolumn{1}{l|}{63.73}  & \multicolumn{1}{l|}{\bftab 64.60}  & \multicolumn{1}{l|}{62.28}  & \multicolumn{1}{l|}{55.16}  & \multicolumn{1}{l|}{60.98}  & \multicolumn{1}{l|}{63.65}  & \multicolumn{1}{c}{/} \\ 
		DBP: Albums &  \multicolumn{1}{l|}{75.24}  & \multicolumn{1}{l|}{79.31}  & \multicolumn{1}{l|}{\bftab 79.99}  & \multicolumn{1}{l|}{54.45}  & \multicolumn{1}{l|}{66.89}  & \multicolumn{1}{l|}{79.38}  & \multicolumn{1}{c}{/} \\ 
		DBP: Movies &  \multicolumn{1}{l|}{80.06}  & \multicolumn{1}{l|}{\bftab 80.48}  & \multicolumn{1}{l|}{78.89}  & \multicolumn{1}{l|}{59.40}  & \multicolumn{1}{l|}{68.11}  & \multicolumn{1}{l|}{78.84}  & \multicolumn{1}{c}{/} \\ \bottomrule
		
	\end{tabular}
	\caption{The accuracy scores obtained by various techniques on different datasets.}
	\label{tab:accuracies}
\end{table*}

\begin{table}
    \centering
	\begin{minipage}{.45\textwidth}
		\flushleft
		\begin{tabular}{|l|c|c|c|} \hline
				Strategy & Pears. $r$ & Spear. $\rho$ & $\mu$ \\ \hline
				Random & \textbf{0.578} & 0.390 & 0.466 \\ \hline
				Anonymous & 0.321 & 0.324 & 0.322 \\ \hline
				Walklets & 0.528 & 0.372 & 0.437 \\ \hline
				HALK  & 0.455 & 0.376 & 0.412 \\ \hline
				N-grams & 0.551 & 0.353 & 0.431 \\ \hline
				WL    & 0.576 & \textbf{0.412} & \textbf{0.480} \\ \hline
		\end{tabular}%
	\end{minipage}
	\begin{minipage}{.2\textwidth}
		\flushright
		\begin{tabular}{|l|c|} \hline
			Strategy & Kendall $\tau$ \\ \hline
			Random & \textbf{0.523} \\ \hline
			Anonymous & 0.243 \\ \hline
			Walklets & 0.520 \\ \hline
			HALK  & 0.424 \\ \hline
			N-grams & 0.483 \\ \hline
			WL    & 0.516 \\ \hline
		\end{tabular}%
	\end{minipage}
	\caption{Document similarity and entity relatedness results}
	\label{tab:results2}
\end{table}

\section{Discussion}\label{sec:discussion}

Based on the provided results, several observations can be made. The \emph{random} and \emph{WL} are used in the original RDF2Vec study~\cite{ristoski2019rdf2vec}. As such, the results reported in this study can be seen as a reproduction of those results. It is important to note here that the only reason why the results obtained by the \emph{WL} and \emph{random} strategy differ in this and the original work, is because walks are extracted after each iteration of the \emph{WL} relabelling algorithm. This results in $k$ times as many walks, with $k$ the number of iterations in the relabelling algorithm. If walks from only one of the iterations would be used, the results would be identical to those of the \emph{random} strategy. Nevertheless, this simple trick does often result in increased predictive performances, as was empirically shown by Ristoski et al.~\cite{ristoski2019rdf2vec}. We hypothesize that this is due to more weight being given, internally in Word2Vec, to the entities where many walks can be extracted from. While the original \emph{WL} and \emph{random} strategies result in very strong performances, especially on the downstream tasks of DBpedia they are often outperformed by custom strategies proposed in this work. \\

While the results indicate that there is no \textit{one-size-fits-all} walking strategy for all tasks and datasets, it seems that the \emph{n-gram} strategy results in the best predictive performances on average for node classification tasks. The average rank of the \emph{n-gram} strategy on the four node classification and three citation network datasets, using all seven techniques, is equal to $1.86$, followed by $3$ of the \emph{HALK} strategy and $3.07$ of both the \emph{random} and \emph{WL} strategy. An average rank of $1$ would mean that the technique outperforms all others on each dataset. The average rank of the \emph{n-gram} strategy on all the node classification tasks, excluding the \emph{community} strategy, is equal to $2.08$, followed by $2.375$, $2.875$ and $3.67$ by \emph{random}, \emph{WL} and \emph{HALK} respectively. \\


The performance of the \emph{community} strategy varies a lot. On some datasets, such as AIFB and BGS, its performance is among the best while it performs a lot worse than \emph{random} walks on others. This is due to the fact that the quality of the walks is highly dependent on the quality of the community detection. If the groups of nodes, clustered by the community detection, do not align well with the downstream task, the performance worsens. \\

Further, it is important to note that the various strategies are complementary to each other. Even when equal accuracies are achieved, the confusion matrices can differ. Therefore, the combination of several strategies can further increase the predictive performance. There are different points within the pipeline where the combination of strategies can take place: (i) at corpus level before feeding the walks to Word2Vec, (ii) at embedding level, by combining the different produced embeddings, and (iii) at prediction level, by aggregating the predictions of the different models. We consider this combination of strategies to be an interesting future step. While the predictive performances of some of the strategies proposed in this work, such as the \emph{anonymous} and \emph{walklet} strategy, do often not come near that of the \emph{random} strategy, a combination of these strategies could improve performance. \\

Some limitations of this study can be identified. Firstly, no comparisons with other techniques are performed. Here, it is important to note that RDF2Vec is an unsupervised and task-agnostic technique. As such, comparisons with supervised techniques, specifically trained for certain tasks, such as Relational Graph Convolutional Networks~\cite{schlichtkrull2018modeling} are rather unfair. In the original work of Ristoski et al.~\cite{ristoski2019rdf2vec} it was already shown that RDF2Vec outperforms other unsupervised variants such as TransE, TransH and TransR. This was independently confirmed by Zouaq and Martel, who additionally showed that RDF2Vec outperformed ComplEx and DistMult as well~\cite{zouaq2020schema}. Second, a fixed depth and fixed hyper-parameters for the Word2Vec model were used within this study. While tuning these hyper-parameters could possibly result in increased predictive performances, it should be noted that the number of hyper-parameters and the range of a Word2Vec model are very large and that the time required to generate the embedding is significant. We therefore opted to fix the hyper-parameters on sensibly chosen values, as was done by Ristoski et al.

\section{Conclusion}\label{sec:conclusion}

In this work, five walk strategies that can serve as an alternative to the basic random walk approach are proposed as a response to the observation that the WL kernel offers little improvement in the context of a single KG. Results indicate that there is no \textit{one-size-fits-all} strategy for all datasets and tasks, and that tuning the strategy to a specific objective, as opposed to simply using the random walk approach, can result in increased predictive performances. \\

There are several future directions that we deem interesting. First, it would be interesting to study the impact on the performance when the strategies are combined with different biased walk strategies and embedding algorithms that differ from the Word2Vec model used within this work. Second, all of the strategies proposed in this work are unsupervised, but supervised approaches could be evaluated that sacrifice generality to gain predictive performance. Third, as already mentioned, the walking strategies are complementary to each other and combining them could potentially result in increased predictive performances. Therefore an evaluation of different combination strategies would be an interesting addition.

\section*{Reproducibility and code availability}
We provide a Python implementation of RDF2Vec with can be combined with any of the walking strategies discussed in this work\footnote{\url{github.com/IBCNServices/pyRDF2Vec}}. Moreover, we provide all code required to reproduce the reported results\footnote{\url{github.com/GillesVandewiele/WalkExperiments}}.

\section*{Acknowledgements}
GV (1S31417N) and BS (1SA0219N) are funded by a strategic base research grant of the Fund for Scientific Research Flanders (\textsc{fwo}).

\bibliographystyle{unsrt}
\bibliography{bibliography.bib}
\end{document}